# A study of retrieval algorithms of sparse messages in networks of neural cliques

Ala Aboudib[*], Vincent Gripon, Xiaoran Jiang[**]

Electronics Department, Télécom Bretagne (Institut Mines-Télécom), Technopôle Brest-Iroise,
CS 83818 – 29238 Brest cedex 3, France

*Abstract*--Associative memories are data structures addressed using part of the content rather than an index. They offer good fault reliability and biological plausibility. Among different families of associative memories, sparse ones are known to offer the best efficiency (ratio of the amount of bits stored to that of bits used by the network itself). Their retrieval process performance has been shown to benefit from the use of iterations. However classical algorithms require having prior knowledge about the data to retrieve such as the number of nonzero symbols. We introduce several families of algorithms to enhance the retrieval process performance in recently proposed sparse associative memories based on binary neural networks. We show that these algorithms provide better performance, along with better plausibility than existing techniques. We also analyze the required number of iterations and derive corresponding curves.

*Keywords*--associative memory, sparse coding, artificial neural network, parsimony, clique, distributed algorithms, iterative processing.

## I. INTRODUCTION

A new artificial neural network model was proposed recently by Gripon and Berrou [1]. This model employs principles from the information theory and error correcting codes and aim at explaining the long-term associative memory functionality of the neocortex. The original model introduced in [1] was proved to outperform the celebrated Hopfield neural network [2] in terms of diversity (the number of messages the network can store) and efficiency (the ratio of the amount of useful bits stored to that of bits used to represent the network itself) [3]. In fact this model can be seen as a particular Willshaw sparse associative memory [4] [5] with constraints on the structure that enable improved retrieval algorithms [6]. Despite these constraints, the model offers asymptotically equivalent performance as generic Willshaw networks. In [7], the authors propose improved retrieval algorithms for the original model of [1].

In this paper, we modify the model proposed by Gripon and Berrou in order to provide their networks with more flexibility. We obtain a model similar to the original one proposed by Willshaw while keeping the interesting features of the structure and thus keeping enhanced retrieval algorithms. We introduce a generic formulation of these algorithms, that contain those proposed by Willshaw [4], Palm [8] and Gripon and Berrou [1]

[*]The authors are also with the Laboratory for Science and Technologies of Information, Communication and Knowledge, CNRS Lab-STICC, Brest, France. Tel.: +33770322057, Email addresses: ala.aboudib@telecom-bretagne.eu, Vincent.gripon@telecom-bretagne.eu, xiaoran.jiang@telecom-bretagne.eu.

[**]This work was partially supported by the European Research Council (ERC-AdG2011 290901 NEUCOD).

along with new ones providing better performance.

The rest of this paper is organized in seven parts; in section II we describe the general architecture of the network by Gripon and Berrou and explain how it can be extended. Section III introduces a generic formulation of the retrieval algorithm. Then, the following four sections develop each step of this algorithm. For each step, different techniques are reviewed if available. Some of these techniques have been proposed in previous papers, and others we introduce here for the first time. In section XIII, performance comparisons of several combinations of retrieval techniques are presented. Section IX is a conclusion.

## II. NETWORK TOPOLOGY AND STORING MESSAGES

This section focuses on the neural-network-based auto-associative memory introduced by Gripon and Berrou in [1]. It is dedicated to defining this network and describing how it can be extended to store variable-length messages. In part *A* of this section the raw structure of the network is described. That is, the network structure before any message is stored. After that, we describe in part *B* how data are structured (stored) within the network in such a way that associative memory functionality emerges.

*a) Raw Structure*

The network can be viewed as a graph consisting of $n$ vertices initially not connected (zero adjacency matrix) organized in $\chi$ parts called clusters with each vertex belonging to one and only one cluster. Clusters are not necessarily of equal sizes but for simplicity, they will be all considered of size $l$ throughout this work. Each cluster is given a unique integer label between 1 and $\chi$, and within each cluster, every vertex is given a unique label between 1 and $l$. Following from this, each vertex in the network can be referred to by a pair $(i,j)$, where $i$ is its cluster label, and $j$ is the vertex label within cluster $i$. For biologically-inspired reasons, the authors of [1] use the term fanal instead of vertex.

At a given moment $t$, a binary state $v_{ij}$ is associated with each fanal $(i,j)$ in the network. $v_{ij}$ is given the value 1 if $(i,j)$ is active and 0 if it is idle. Initially, all fanals are supposed to be idle.

The adjacency matrix (also called the weight matrix $w$) for this graph is a binary symmetric square matrix whose elements take values in $\{0,1\}$ where a zero means an absence of a connection while a 1 indicates that a connection is present. Line and column indices are fanals represented as $(i,j)$ pairs. So in order to indicated that two fanals $(i,j)$ and $(i,j)$ are connected, we write $w_{(i,j)(i,j)} = 1$. All connection combinations are allowed except those among fanals belonging to the same cluster, resulting in a $\chi$-partite graph. When the memory is empty, $w$ is a zero matrix.

*b) Message storing procedure*

This network is designed to store sparse messages of the form $[|0,l|]^\chi$. So each message consists of $\chi$ integer sub-messages or segments in $[|0,l|]$ where a 0 refers to a segment that needs not to be stored. We are only interested in storing non-zero sub-messages. There are $c$ of them. For example, in order to store a message $m = \{0,10,7,0,12,11\}$ in a network with $\chi = 6$ and $l = 12$, each non-zero segment position $i$ within this message is interpreted as a cluster label, and the segment value $j$ is interpreted as a fanal label within the cluster $i$. Thus, each non-zero segment is associated with a unique fanal $(i,j)$. Thus, this example maps to the 10th fanal of the 2nd cluster, the 7th fanal of the 3rd cluster, the 12th fanal of the 5th cluster and the 11th fanal of the 6th (last) cluster.

Then, given these $c$ elected fanals in $c$ distinct clusters, the weight matrix of the network is updated according to equation (1) so that a fully connected sub-graph called a clique is formed of these elected fanals.

$$w_{(i,j)(\hat{i},\hat{j})} = 1 \quad if \ i \neq \hat{i} \ and \ m_i, m_{\hat{i}} \neq 0 \quad (1)$$

where $(i,j)$ and $(\hat{i},\hat{j})$ are fanals associated to message segments $m_i$ and $m_{\hat{i}}$ respectively.

It is important to note that if we wish to store another message $\acute{m}$ that overlaps with $m$. That is, $\acute{m}$'s corresponding clique shares one or more connections with that of $m$, the value of these connections, which is 1, should not be modified, these connections are simply reused. So the storing process is non-destructive and the network's connection map is the union of all cliques. Figure 1 depicts an example of a network with $n = 72$, $\chi = 6$ storing two messages of size $c = 4$.

It is worth noting that when $l = 1$, the structure of this network becomes equivalent to the Willshaw model.

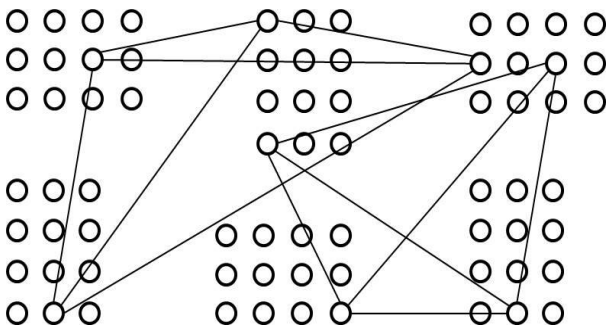

**Figure 1** Learning sparse messages. Each message is stored as a clique of fanals belonging to a subset of the overall set of clusters.

## III. THE RETRIEVAL PROCESS

The goal of the retrieval process introduced in this paper is to recover an already stored message (by finding its corresponding clique) from an input message that has undergone partial erasure. A message is erased partially by replacing some of its non-zero segments by zeros. For example, if $A = \{0,1,8,0,10,12\}$ is a message stored in the network of figure 1, we can imagine $\bar{A}_{erased} = \{0,0,0,0,10,12\}$.

The core of the retrieval process can be viewed as an iterative twofold procedure composed of a dynamic rule and an activation rule. Algorithm 1 depicts the steps of the retrieval process:

**Algorithm 1** The retrieval process.

*Insert an Input Message.*
*Apply a dynamic rule.*

*Phase 1:*
  *Apply an activation rule.*
  *Apply a dynamic rule.*

*Phase 2:*
  *While (stopping criterion is not attained) {*
    *Apply an activation rule.*
    *Apply a dynamic rule.*

  *}.*

*output ← active fanals.*

Each step of algorithm 1 is described in detail in the next sections.

## IV. INPUT MESSAGE INSERTION

An input message should be fed to the network in order to trigger the retrieval process. For example, suppose that we have a message $A = \{7,1,5,11,0,0\}$ stored in the network of figure 1 where $\chi = 6, c = 4, l = 12$. Suppose now that we wish to retrieve $A$ from the partially erased input $\bar{A} = \{0,0,5,11,0,0\}$. In order to do that, all fanals corresponding to non-zero segments should be activated. That is, a fanal $(i,j)$ associated with the segment of $\bar{A}$ at position $i$ whose value is $j$ is activated by setting $v_{ij} = 1$. So, $\bar{A}$ activates two fanals: (3,5) and (4,11).

It is worth noting that from the point of view of the network, erased segments and segments corresponding to irrelevant clusters i.e. clusters exempted from storing a message, are all seen as zeros and thus they are initially indistinguishable. So, the job of the retrieval process is to determine which clusters are relevant and then to find the fanals within these clusters that represent the searched message.

Now, having a number of active fanals, a dynamic rule should be applied.

## V. DYNAMIC RULES

A dynamic rule is defined as the rule according to which fanal scores are calculated. We will denote the score of a fanal $(i,j)$ by $\lambda_{ij}$. Calculating fanals' scores is crucial to deciding which ones will be activated. A score is a way to estimate the chance that a fanal belongs to a bigger clique within the set of active fanals and thus the chance that it belongs to the message we are trying to recover. In principle, the higher the score the higher this chance is. Two dynamic rules have been already introduced, namely, the SUM-OF-SUM [1] and the SUM-OF-MAX [6] rules. We also introduce for the first time what we call the NORMALIZATION rule.

*a) The SUM-OF-SUM rule (SOS)*

The SUM-OF-SUM is the original rule. It states that the score of a fanal $(i,j)$ denoted by $\lambda_{ij}$ is simply the number of active fanals connected to $(i,j)$ plus a predefined memory effect $\gamma v_{ij}$. Scores should be calculated for all of the fanals in the network. This SUM-OF-SUM rule can be formalized by the following equation:

$\forall i \ and \ j, 1 \leq i \leq \chi, 1 \leq j \leq l$:

$$\lambda_{ij} = \gamma v_{ij} + \sum_{\hat{i}=1}^{\chi}\sum_{\hat{j}=1}^{l} w_{(ij)(\hat{i}\hat{j})} v_{\hat{i}\hat{j}}. \quad (2)$$

Although intuitive, this rule has a major problem which is that in some cases, the scores give a false estimate of the chance that a given fanal belongs to a bigger clique within the set of active fanals. To clarify this point we consider the example of figure 2 where black circles represent active fanals and gray ones represent fanals for which we are going to calculate the scores. According to the SUM-OF-SUM rule, the fanal in the cluster on top-left has a score of 3 while the fanal in the cluster on bottom-right has a score of 4. This result indicates that the latter fanal is more likely to belong to a bigger clique than the former because it has a higher score. This observation is not true since all the active fanals that contribute to that higher score belong to the same cluster and thus no connections are allowed to form among them.

In order to solve this problem the SUM-OF-MAX and the NORMALIZATION dynamic rules can be applied.

*b) The NORMALIZATION rule (NORM)*

In the NORMALIZATION rule that we introduce here, fanals' scores are calculated using the following equation:

$$\lambda_{ij} = \gamma v_{ij} + \sum_{i=1}^{\chi} \frac{1}{|v_i|} \sum_{j=1}^{l} w_{(ij)(ij)} v_{ij}. \qquad (3)$$

where $|v_i|$ is the number of active fanals in cluster $i$. Equation (3) states that the contribution of a fanal $(i,j)$ to the score of another fanal connected to it is normalized on the number of active fanals in cluster $i$. That is, if the cluster $i$ contains $x$ active fanals, then the contribution of the fanal $(i,j)$ becomes $1/x$. So by applying this rule to the network of figure 2, the fanal in the cluster on bottom-right gets a score of 2 and the fanal on top-left gets a score of $2\frac{1}{3}$ which is a result that favorises the latter fanal to be activated and thus solves the SUM-OF-SUM rule problem.

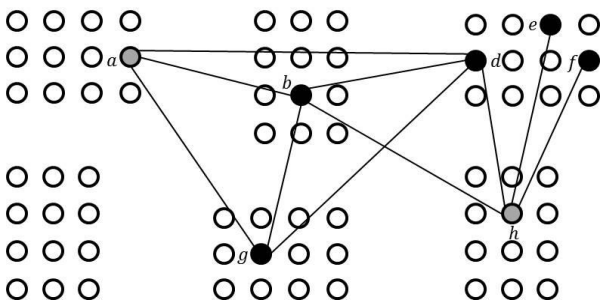

**Figure 2** Dynamic rule application phase. Black-filled fanals are active fanals and gray-filled ones are fanal to which scores are to be calculated.

*c) The SUM-OF-MAX rule (SOM)*

According to the SUM-OF_MAX rule, the score of a fanal $(i,j)$ is the number of clusters in which there is at least one active fanal $(i,j)$ connected to $(i,j)$ plus the memory effect $\gamma v_{ij}$:

$$\lambda_{ij} = \gamma v_{ij} + \sum_{i=1}^{\chi} \max_{1 \le j \le l}(w_{(ij)(ij)} v_{ij}). \qquad (4)$$

So referring back to figure 2, and according to equation (4), the fanal on top-left has a score of 3 where the other fanal on bottom-right has a score of 2. This is a more satisfying result than the one obtained by the SUM-OF-SUM rule since it indicates that the latter fanal, although connected to more active fanals, is less likely to belong to a bigger clique within the set of active fanals than the one on top-left.

Moreover, it has been shown in [6] that for the particular case, when $c = \chi$, the SUM-OF-MAX rule guarantees that the retrieved massage is always either correct or ambiguous but not wrong. An ambiguous output message means that in some clusters more than one fanal might be activated of which one is the correct fanal.

VI. ACTIVATION RULES

The activation rule is applied for electing the fanals to be activated based on their scores after application of a dynamic rule. So basically, a fanal $(i,j)$ is activated if its score $\lambda_{ij}$ satisfies two conditions:
- $\lambda_{ij}$ attains a certain value defined by each activation rule.
- $\lambda_{ij} \ge \sigma_{ij}$ where $\sigma_{ij}$ is the activation threshold of fanal $(i,j)$. [9]

The activation rule should be able to find two unknowns: The subset of clusters to which the message we are trying to recover belongs, and the exact fanals within these clusters representing the sub-messages. Two activation rules are introduced in this paper: the Global Winners Take All rule (GWsTA) which is a generalization of the Global Winner Take All (GWTA) rule, and the Global Losers Kicked Out (GLsKO) rule which is a new rule we propose here for the first time.

*a) The GWsTA rule*

The GWTA rule introduced in [10] activates only fanals with the network-wide maximal score. The problem with this rule is that it supposes that fanals belonging to the message we are looking for have equal scores. It also supposes that this unified score is maximal which is not necessarily the case. It has been shown in [10] that spurious cliques, i.e. cliques that share one or more edges with the clique we are searching, might appear and render the scores of the shared fanals of the searched clique higher than others'.

For example, in the network of figure 2, if the searched clique is $abdg$, then $bdh$ is a spurious one. Now by applying the SUM-OF-MAX rule on the black and gray fanals which are all supposed active, and considering $\gamma = 1$ and that the network contains only the connections shown in the figure, we get the scores: $a: 4$, $b: 5$, $d: 5$, $e: 2$, $f: 2$, $g: 4$, $h: 3$. Thus, according to the GWTA rule, only fanals $b$ and $d$ will be kept active and the clique $abdg$ is lost. This is caused by the spurious clique $bdh$ which highers the scores of $b$ and $d$.

The generalization of the GWTA rule we propose is meant to account for this problem.

The behavior of the Global Winners Take All rule is the same in both phases of the retrieval process. It elects a subset of fanals with maximal and near-maximal scores to be activated. In other words, it defines a score threshold $\theta$ at each iteration, and only fanals that have at least this score threshold are activated.

In order to calculate this score threshold at a given iteration, we first fix an integer parameter $\alpha$. Then we make a list compromising the $\alpha$ highest scores in the network including scores that appear more than once. For example, if the fanals scores in a network with $n = 10$ are $\{25, 18, 25, 23, 23, 19, 18, 19, 18, 17\}$ and $\alpha = 7$, then our list becomes $\{25, 25, 23, 23, 19, 19, 18\}$. The minimum score in this list which is 18 becomes the score threshold $\theta$. Then we apply the equation:

$\forall i$ and $j$, $1 \leq i \leq \chi$, $1 \leq j \leq l$:

$$v_{ij} = \begin{cases} 1, & \lambda_{ij} \geq \theta \text{ and } \theta \geq \sigma_{ij} \\ 0, & \text{otherwise} \end{cases} \quad (5)$$

It is worth noting that this activation rule is equivalent to the retrieval rule proposed by Willshaw [4] in that fanals are activated by comparing their scores to a fixed threshold $\theta$. The basic differences is that Willshaw does not specify a rule for choosing $\theta$ and that his retrieval process is not iterative (only one iteration). In this context, the GWsTA rule can be considered as a rule for fixing an optimal value of $\theta$ using the method described above where $\alpha = c$ optimally.

One problem with this rule is that the choice of an optimal $\alpha$ for a certain message size would not be adapted for other message sizes. This limits the possibility of using this rule for retrieving messages of variable sizes. Moreover, this rule always activates a subset of fanals with maximal and near-maximal scores. But in some cases, when the number of stored messages reaches a high level, the fanals with the maximal score does not necessarily belong to the searched message.

*b) The GLsKO rule*

The Global Losers Kicked Out (GLsKO) has a behavior in *phase 1* of the retrieval process that differs from that of *phase 2* as follows:

---

**Phase 1:**
Apply the GWTA rule.

**Phase 2:**
Kick losers out.

---

In *phase 1*, the GWTA rule is applied resulting in the activation of a subset of fanals to which the searched message is guaranteed to belong. After this, the activation thresholds of inactive fanals are set to infinity because we are no more interested in searching outside the already activated fanals.

In *phase 2*, the rule changes behavior, so, at each iteration, it makes a list compromising the $\beta$ lowest non-zero scores of the active fanals only. For example, if the set $\{25, 18, 25, 23, 23, 19, 18, 19, 17, 17\}$ represents the scores of active fanals in a network with $n = 10$ and we fix $\beta = 3$, then the list of lowest scores becomes $\{18, 19, 18, 19, 17, 17\}$. After that, a score threshold $\theta$ equals to the maximum score in the latter list is set, and only fanals with scores greater than $\theta$ are kept active. This can be described by the following equation:

$\forall i$ and $j$, $1 \leq i \leq \chi$, $1 \leq j \leq l$:

$$v_{ij} = \begin{cases} 1, & \lambda_{ij} > \theta \text{ and } \theta \geq \sigma_{ij} \\ 0 \text{ and } \sigma_{ij} \to \infty, & \text{otherwise} \end{cases} \quad (6)$$

An optional property of the GLsKO rule is the ability to determine the number of fanals $\mu$ to be deactivated. Which we found to be only interesting when $\beta = 1$. For example, if we set $\beta = 1$ and $\mu = 1$ in the network example of the previous paragraph, we get the following list of scores $\{17,17\}$. If $\mu$ is not specified, all losers are deactivated. But since $\mu = 1$, only one of these two fanals is randomly chosen to be deactivated. This may be useful if we wish to exclude losers one at a time and thus reduce incautious quick decisions. Another property of this rule is that the choice of $\beta$ is independent of the message sizes. This should allow this rule to retrieve stored messages even if they differ in size which is not the case with the GWsTA as we have seen.

## VII. STOPPING CRITERIONS

Since the retrieval process is iterative, a stopping criterion should be used in order to put this process to an end. In parts *A* and *B* of this section we review the two classic criterions that are already in use. In parts *C* and *D* we propose two new ones that are supposed to enhance performance.

*a) A fixed number of iterations (ITER)*

A stopping criterion can be defined as a predefined number of iterations of the retrieval process. So dynamic and activation rules are applied iteratively, and when a counter announces that the desired number of iterations is attained, the retrieval process terminates and the activated fanals are taken for the retrieved message. The problem with this approach is that the stopping criterion which is a simple iteration counter is independent of the nature of retrieved message. That is, the activated fanals after the last iteration are not guaranteed to form a clique corresponding to an already stored message. This stopping criterion is only interesting with the GWsTA rule.

*b) The convergence criterion (CONV)*

This criterion states that if the set of active fanals at iteration $t + 1$ is the same as that of iteration $t$, the retrieval process is taken as converged so it terminates and the result is output. The convergence criterion is only compatible with the GWsTA rule. In the case of the GLsKO rule, one or more active fanals are deactivated in each iteration. So it is not possible to have the same set of active fanals throughout two subsequent iterations.

*c) The equal scores criterion (EQSC)*

The idea we propose for this criterion is that when a point is reached where there are no more losers. Or, in other words, when all scores of active fanals are equal, the retrieval process terminates and the result is output.

*d) The clique criterion (CLQ)*

This new criterion depends on the relationship between the number of activate fanals and their scores. If the activate fanals form a clique the retrieval process terminates. Thus the retrieved message is more likely to make sense though it is not necessarily the correct result. In order to check if the activated fanals form a clique, we define $V_a$ as the set of active fanals and $\lambda(v_{ai})$ as the score of the active fanal $v_{ai}$, then we apply the following procedure:

---

$\forall\ 1 \leq i,j \leq |V_a|$.
If
$\quad \lambda(v_{ai}) = \lambda(v_{aj}) = \rho \quad$ and $\quad |V_a| = \rho - (\gamma - 1)$.
Then
$\quad output\ the\ result$.

*terminate the retrieval process*.

---

To make sense of this stopping criterion, we take an intuitive situation when $\gamma = 1$. In this case, the stopping criterion is that when all active fanals have an equal score which is equal to the number of these fanals, a clique is recognized, so the process is terminated and the result is output.

It is worth noting that when using the GWsTA rule, it is always preferable to combine any stopping criterion with the ITER criterion to prevent infinite looping since this rule is not guaranteed to always satisfy other stopping criterions.

## VIII. RESULTS

We have seen that there are many possible combinations of dynamic, activation rules and stopping criterions in order to construct a retrieval algorithm. In this section we will demonstrate the performance of some of these combinations.

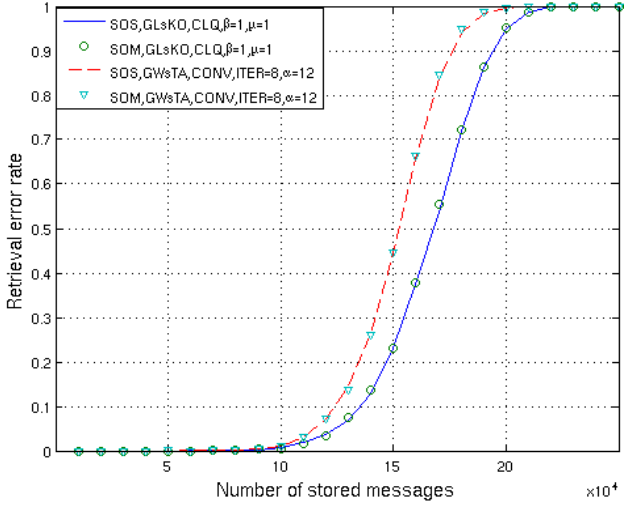

**Figure 3** Influence of dynamic rules on retrieval error rates in a network with $\chi = 100, l = 64, c = 12, \gamma = 1, \sigma_{ij} = 0$ initially, with 3 segments of partial erasure in input messages.

*a) First observation*

Figure 3 shows that both the SOM and the SOS dynamic rules give the same performance. The NORM rule was found to give the same results also, but it is not shown in the figure for clarity. This is not the case with the original network introduced in [1] where the SUM-OF-MAX rule was proved to give better results [6]. This is an interesting phenomenon that is worth studying. It indicates that the major source of retrieval errors networks string sparse messages is not related to the different methods of calculating fanals' scores.

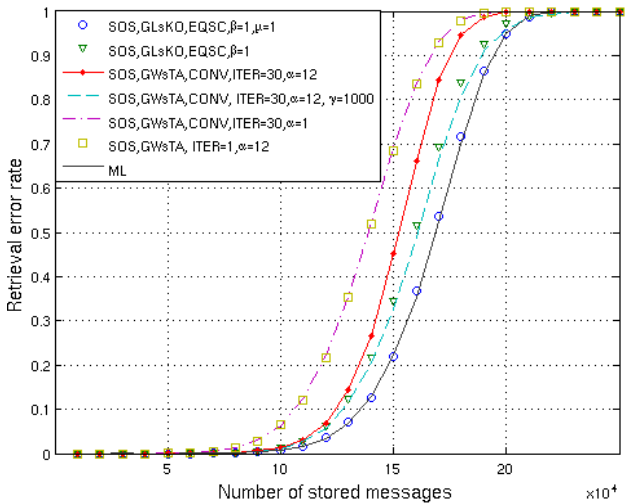

**Figure 4** Influence of activation rules on retrieval error rates in a network with $\chi = 100, l = 64, c = 12, \gamma = 1$ if not stated otherwise, $\sigma_{ij} = 0$ initially, with 3 segments of partial erasure in input messages.

*b) Second observation*

We notice in figure 4 that the GWsTA ($\alpha = 12$) rule gives a better performance than GWTA (GWsTA, $\alpha = 1$) used with the CONV stopping criterion combined with a maximum possible number of iterations of 30. We also notice that the GWsTA ($\alpha = 12$) gives a lower error rate when the memory effect $\gamma = 1000$. However, the GLsKO ($\alpha = 1$, $\mu = 1$) rule using the EQSC or the CLQ (not shown on the figure for clarity) stopping criterion has the lowest error rate which attains the performance of the brute force retrieval algorithm (ML) for this amount of input erasure. But the performance of the GLsKO rule with $\alpha = 1$ and $\mu$ unspecified degrades.

It is also worth pointing out that the GWsTA ($\alpha = 12$) rule with one iteration which is equivalent to the retrieval rule proposed by Willshaw [4] gives a result similar to that of the GWTA rule used with the CONV and ITER = 30 stopping criterions which turns out to be the worst performance. This proves that combining iterations with a proper activation threshold $\theta$ significantly improves performance as shown in [8].

*c) Third observation*

Figure 5 shows that the average number of iterations required to retrieve a message is relatively constant for all rules up to 140000 messages learnt. Beyond this, the number of iterations required for the GLsKO and the GWsTA rules with $\gamma = 1$ begins to increase rapidly. It is worth emphasizing that the maximum number of iterations allowed for the GWsTA rule is 30 so the constant level reached by the curve representing this rule with $\gamma = 1$ in figure 5 is just a result of that constraint. However, the number of iterations for the GWsTA rule with $\gamma = 1000$ increases only slightly approaching an average of 3.3 up to 250000 messages learnt.

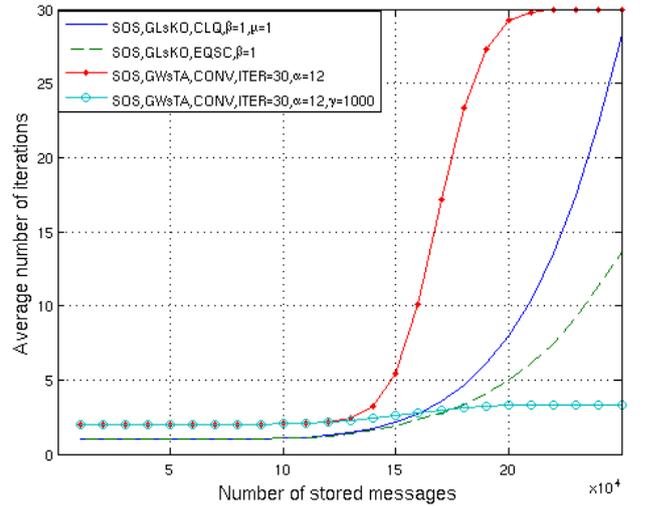

**Figure 5** Average number of iterations for different scenarios in a network with $\chi = 100, l = 64, c = 12, \gamma = 1$ if not stated otherwise, $\sigma_{ij} = 0$ initially, with 3 segments of partial erasure in input messages.

## IX. CONCLUSION

In this paper, we presented all existing retrieval algorithms and suggested new ones for a recently invented sparse associative memory model. We also demonstrated and compared the performance of these algorithms when using partially erased messages as inputs. We aim to extend the scope of the algorithms presented in this paper to deal with other types of input messages such as distorted ones.

We found that the GLsKO activation rule combined with the equal scores or the clique stopping criterions gives the best results in terms of retrieval error rate but with a rapidly increasing number of iterations. Actually, the second phase of the GLsKO rule along with the clique criterion can be viewed as an operation equivalent to searching the maximum clique among active fanals. This is a famous NP-complete problem with no optimal solution yet. However, many sub-optimal solutions were suggested for this problem (or equivalently, the minimum vertex cover problem) such as [11] [12] and many more. We believe that such sub-optimal solutions are adaptable to our problem and can be integrated in our retrieval algorithm in the future in order to give a better performance with a more reasonable number of iterations.

Finally, the retrieval algorithms presented in this work are all synchronous in the sense that, at each iteration, dynamic and activation rules are always applied to all clusters. In future work, we will consider asynchronous methods which can take into account the fact that some clusters may reach their final state before others, so application of dynamic and activation rules could then be limited to only a subset of clusters.

## ACKNOWLEDGMENT

The authors would like to thank Zhe Yao and Michael Rabbat at McGill University for interesting discussions that helped leading to the proposal of the new retrieval techniques introduced in this paper. We also acknowledge funding from the NEUCOD project at the electronics department of Télécom Bretagne.